\documentclass[letterpaper]{article} 
\usepackage{aaai2026}  
\nocopyright
\usepackage{times}  
\usepackage{helvet}  
\usepackage{courier}  
\usepackage[hyphens]{url}  
\usepackage{graphicx} 
\usepackage{pgfplots}
\usepackage{pgfplots}
\usepackage{sansmath}
\pgfplotsset{compat=1.18} 
\usepackage{natbib}  
\usepackage{caption} 
\usepackage{tabularx}

\usepackage{algorithm}
\usepackage{algorithmic}
\usepackage{booktabs}
\usepackage{multirow}
\usepackage{amsmath} 
\usepackage{amssymb}
\usepackage{amsfonts}
\usepackage{nicematrix}
\usepackage{subcaption}
\usepackage{adjustbox}
\usepackage{makecell}
\usepackage{xcolor}

\usepackage{colortbl}
\definecolor{mygray}{gray}{0.9}

\usepackage{newfloat}
\usepackage{listings}
\DeclareCaptionStyle{ruled}{labelfont=normalfont,labelsep=colon,strut=off} 
\floatstyle{ruled}
\newfloat{listing}{tb}{lst}{}
\floatname{listing}{Listing}
\title{LongDPM: Overlap-Aware 4D Reconstruction from Long Monocular Videos}

\author{
    Chenyi Xu,
    Yihao Wu,
    Liqi Yan\thanks{Corresponding authors.},
    Chao Yang,
    Jianhui Zhang,
    Fangli Guan,
    Pan Li
}
\affiliations{
    Hangzhou Dianzi University
}

\usepackage{bibentry}

\begin{document}

\maketitle

\begin{abstract}
Recovering a dynamic 3D scene from a long monocular video is crucial for dense geometry, camera motion, and temporal correspondence to remain consistent in a shared coordinate system. Existing methods face two key challenges: (1) feed-forward reconstruction models provide accurate local predictions but are limited to short clips, and (2) long-range trackers preserve correspondences without producing dense sequence-level reconstruction. This paper presents \textbf{LongDPM}, a novel overlap-aware framework for scalable long-range monocular dynamic reconstruction. \textbf{First}, LongDPM processes long videos in overlapping chunks, keeping inference memory bounded by the chunk length. \textbf{Second}, it connects chunk-local coordinate systems through confidence-weighted registration with static-aware overlap abstraction. \textbf{Third}, it associates dynamic identities across chunk boundaries and fuses matched trajectories to recover coherent long-range 3D motion. Experimental results demonstrate that LongDPM achieves superior long-range reconstruction and tracking performance, reducing dense tracking EPE over V-DPM on PointOdyssey, Kubric-F, and Kubric-G, while obtaining the best TUM-dynamics ATE for camera pose estimation.

\end{abstract}

\section{Introduction}

\begin{figure*}[!t]
    \centering
    \includegraphics[width=\textwidth]{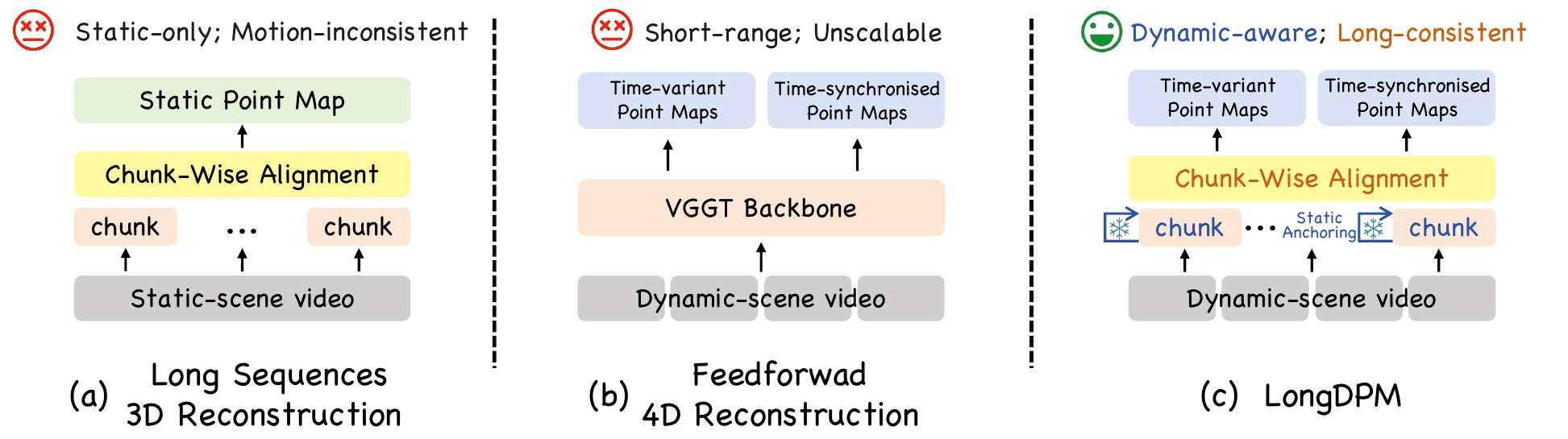}
    \caption{
    \textbf{LongDPM for long-range dynamic reconstruction.}
    (a) Chunk-wise long-sequence 3D reconstruction works well for static videos, but fails to maintain motion consistency in dynamic scenes.
    (b) Short-context 4D reconstructors recover dynamic point maps within local clips, yet are difficult to scale directly to long videos.
    (c) LongDPM bridges this gap with dynamic-aware chunk-wise alignment, enabling long-range consistent 4D reconstruction from long monocular dynamic videos.
    }
    \label{fig:teaser}
\end{figure*}

Tracking dynamic points in monocular video is a fundamental capability for understanding scene motion, object interaction, and camera behavior in 3D. Long-sequence tracking is especially important for realistic applications such as mobile capture, robotics, and embodied perception, where videos often span hundreds or thousands of frames rather than short isolated clips. In this setting, the task is to recover persistent correspondences and 3D trajectories for visible scene points while placing them, together with dense geometry and camera motion, in a shared coordinate system over the entire video. This remains challenging because long videos contain large viewpoint changes, independently moving objects, repeated occlusions, exits, and re-entries, so local geometric predictions and short-range correspondences can easily become inconsistent as temporal distance grows.

Recent progress has supplied several ingredients for this goal, but they have largely been developed in separate lines of work. 
Feed-forward 3D reconstruction has shown that dense pointmaps, camera poses, and cross-view correspondences can be inferred directly from images, yielding a compact geometric representation that avoids many hand-crafted stages of classical pipelines~\cite{wang2024dust3r,leroy2024grounding,wang2025vggt}. 
Short-window dynamic reconstruction (or Local 4D reconstruction) extends this capability to monocular videos by predicting time-varying 3D structure and scene flow within short clips, thereby making dynamic point trajectories an integral part of the reconstruction rather than a separate post-processing step~\cite{zhang2024monst3r,sucar2025dynamic,sucar2026v,liang2024feed,zhang2025efficiently,feng2025st4rtrack}. 
Long-sequence reconstruction explores how to scale these local predictions to longer videos, either through streaming states or by aligning overlapping windows into a globally consistent scene model~\cite{chen2025ttt3r,wang2025continuous,zhuo2025streaming,deng2025vggt,xie2026scal3r,xu2026pas3r, chen2025long3r}. 
Long-range point tracking provides complementary temporal cues by preserving point identities across occlusions, deformations, and re-entries—information that is often missing from purely geometric approaches~\cite{doersch2023tapir,karaev2025cotracker3,rajivc2025multi}.

Despite these advances, existing methods still fail in characteristic ways on long dynamic monocular videos~\cite{feng2025st4rtrack,zhang2025efficiently}. \textbf{\textcircled{\scriptsize{1}}} 4D reconstructors designed for local temporal windows produce accurate geometry and motion within each clip, but their fixed context length makes full-sequence inference impractical due to rapidly growing memory and computation costs. Truncation or sparse sampling can reduce this burden, yet discards the dense temporal cues essential for recovering fast motion, occlusion boundaries, and smooth 3D trajectories. \textbf{\textcircled{\scriptsize{2}}} Streaming and chunk-and-align methods extend temporal coverage, but maintaining sequence-level consistency remains challenging. In long dynamic scenes, accumulated pose drift, memory decay, weak parallax, and overlaps dominated by moving objects often cause adjacent chunks to drift apart or align to non-rigid foreground motion rather than the static background. \textbf{\textcircled{\scriptsize{3}}} Long-range tracking methods preserve point correspondences over extended horizons, but are not designed to jointly recover dense scene geometry, camera poses, and trajectories within a single shared 3D coordinate frame~\cite{liu2025trace,wang2025shape}. Moreover, their query-based assumptions can break down in open-world long videos, where points may appear late, disappear for many frames, re-enter under different viewpoints, or lack reliable support near chunk boundaries.

We introduce \textbf{LongDPM}, a framework for long-range monocular dynamic reconstruction. LongDPM builds upon local dynamic predictors to achieve sequence-level consistency through cross-chunk reasoning. It employs a frozen local reconstructor as the short-window predictor and decouples local prediction from global consistency. Specifically, we propose (i)~\emph{Static-Aware Overlap Abstraction} that converts chunk overlaps into confidence-weighted and rigidity-filtered constraints, (ii)~\emph{geometry-conditioned cross-chunk identity association} that links local tracklets in the aligned 3D space using trajectory shape, velocity, and direction cues, and (iii)~\emph{trajectory-guided dynamic fusion} that feeds associated identities back into overlap refinement and boundary reconstruction. The resulting system produces a dense, geometrically consistent 4D reconstruction together with coherent long-range 3D trajectories in a unified coordinate frame, while keeping inference memory bounded by the chunk length.

Our contributions are as follows:
\begin{itemize}
    \item We propose \textbf{LongDPM}, a predictor-agnostic framework that enables long-range 4D reconstruction from monocular videos by performing overlap-aware cross-chunk reasoning on top of local dynamic predictors.
    \item We introduce a static-aware overlap abstraction strategy that decomposes chunk overlaps into static alignment anchors and dynamic supports, distilling noisy overlap regions into confidence-weighted and rigidity-filtered geometric constraints for robust registration in highly dynamic scenes.
    \item We develop a geometry-conditioned identity association module that reconnects chunk-local dynamic tracklets in the aligned 3D space by leveraging trajectory shape, velocity, and direction cues, thereby achieving cross-chunk identity continuity without relying on first-frame queries.
    \item We present a trajectory-guided dynamic fusion procedure that utilizes the associated identities as additional constraints to refine overlap alignment and reconstruct smooth motion transitions across chunk boundaries.
\end{itemize}

\section{Related Work}

\subsection{Feed-forward 3D Reconstruction}
Feed-forward 3D reconstruction has advanced rapidly by replacing scene-specific optimization with direct geometric prediction from images. Pointmap-based methods such as DUSt3R~\cite{wang2024dust3r} estimate dense geometry, correspondences, and camera parameters in a unified representation, avoiding many brittle stages of classical SfM/MVS pipelines~\cite{leroy2024grounding,zhang2025flare,elflein2025light3r}. Building on this paradigm, recent multi-view models jointly infer depth, camera pose, structure, and correspondence from unordered or sparsely sampled image collections~\cite{wang2025vggt,wang2025pi,lin2025depth,keetha2025mapanything,yang2025fast3r}. Extensions to monocular video and dynamic scenes further incorporate temporal cues to recover short-range 4D structure and scene motion~\cite{li2025megasam,zhang2025pomato,zhang2024monst3r,sucar2025dynamic,sucar2026v,han2026enhancing}. Despite these advances, most feed-forward reconstructors are designed around a fixed input context. Their predictions are most reliable within the observed local window, while persistent geometry, camera consistency, and dynamic identities beyond this window are not explicitly modeled.

\subsection{Long-Sequence Feed-forward Reconstruction}
Long-sequence feed-forward reconstruction investigates how to scale local predictors beyond their native temporal support. Streaming and recurrent methods process frames incrementally to improve temporal coverage~\cite{chen2025ttt3r,zhang2026loger,wang2025continuous,zhuo2025streaming,chen2025long3r,chen2026geometric}, but often suffer from error accumulation, memory decay, and sensitivity to early mistakes. Chunk-and-align methods, such as VGGT-Long~\cite{deng2025vggt}, partition videos into overlapping windows and align local reconstructions, thereby scaling feed-forward models to longer RGB sequences~\cite{xie2026scal3r,xiong2026vggt,elflein2026vgg}. However, most existing long-sequence methods primarily target static or camera-centric scenes, where overlap regions are dominated by rigid structures~\cite{deng2025vggt}. In dynamic videos, overlaps frequently contain non-rigid foreground motion, occlusions, low parallax, and unreliable boundary predictions, which can make direct chunk registration unstable. Even when geometric alignment succeeds, dynamic trajectories predicted within each local window often remain fragmented and require additional cross-chunk association.

\subsection{2D/3D Point Tracking for 4D Reconstruction}

Point tracking provides long-range correspondences without directly solving dense 3D reconstruction, which is essential for consistent 4D scene reconstruction. Query-based trackers iteratively refine trajectories from optical flow or learned motion features, enabling them to handle large deformations, fast motion, and temporary occlusions~\cite{teed2020raft,harley2022particle,doersch2023tapir,doersch2024bootstap,tumanyan2024dino}. Joint trackers, such as CoTracker-style models, reason over multiple points jointly via cross-track attention or transformer architectures, improving robustness and supporting quasi-dense tracking~\cite{karaev2024cotracker,karaev2025cotracker3,cho2024local}. Recent 3D trackers further incorporate monocular depth estimates, spatial priors, multi-view constraints, or persistent 3D features to lift tracking from 2D image space into 3D~\cite{xiao2024spatialtracker,xiao2025spatialtrackerv2,zhang2025tapip3d,ngo2024delta,rajic2025mvtracker,alumootil2025dept3r}. Nevertheless, most existing trackers still rely on explicit point queries, typically initialized in the first frame or within a short support window. Moreover, they do not jointly recover dense scene geometry, camera parameters, and trajectories in a single globally consistent 4D coordinate system.

\begin{figure*}[t]
    \centering
    \includegraphics[width=\textwidth]{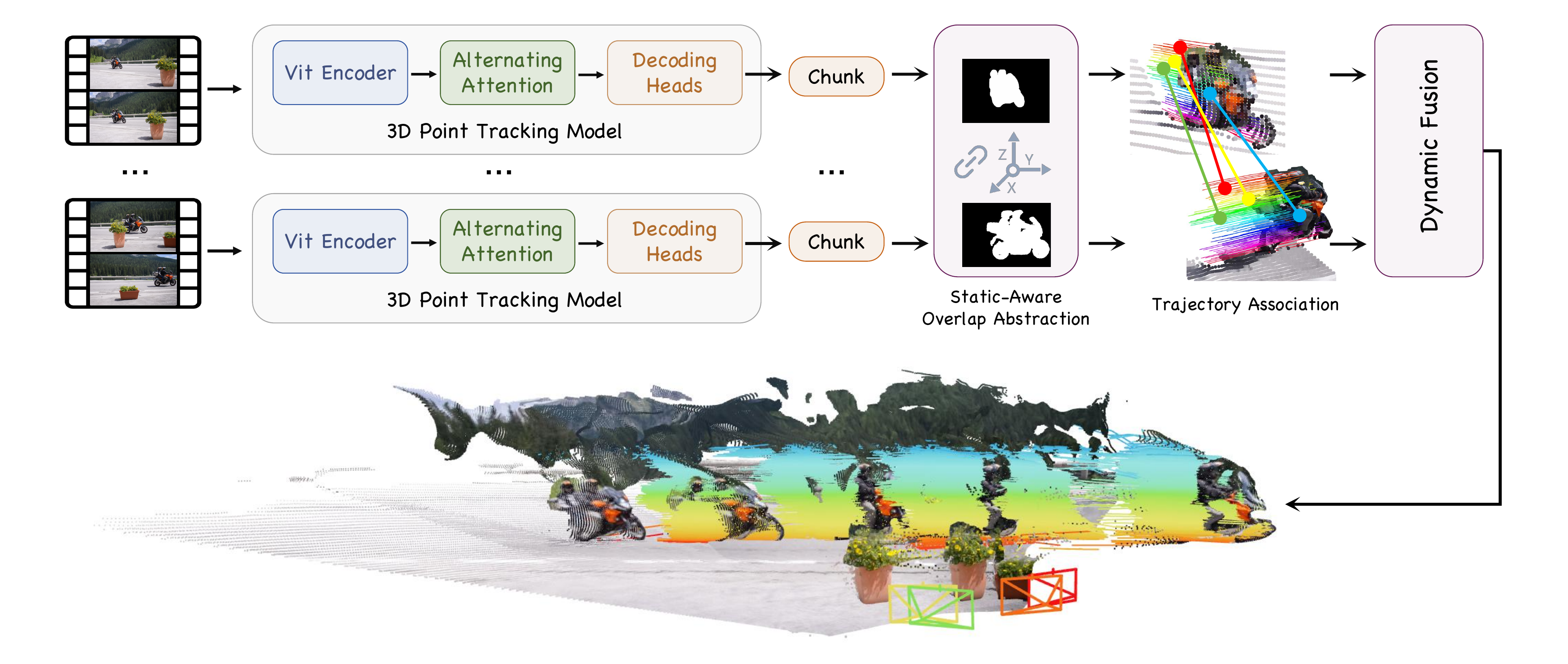}
    \caption{Overview of the LongDPM pipeline. A long monocular video is divided into overlapping chunks for bounded-memory local 4D reconstruction, registered through confidence-weighted overlap constraints, and finally linked across chunk boundaries to recover long-range dynamic trajectories in a shared 3D coordinate system.}
    \label{fig:pipeline}
\end{figure*}

\section{Method}

\subsection{Problem Formulation}
Given a long monocular video $\mathbf{I}=\{I_t\}_{t=0}^{N-1}$, our goal is to recover a sequence-consistent 4D reconstruction in a unified 3D coordinate system. Following the dynamic point map formulation~\cite{sucar2026v}, the output is $\mathcal{R}=\{(\mathbf{P}_t,\mathbf{C}_t,\mathbf{E}_t)\}_{t=0}^{N-1}$, where $\mathbf{P}_t \in \mathbb{R}^{H \times W \times 3}$ is the pointmap, $\mathbf{C}_t \in \mathbb{R}^{H \times W}$ is the confidence map, and $\mathbf{E}_t \in SE(3)$ is the camera pose; the time-varying pointmaps naturally induce dense 3D trajectories $\mathcal{T}=\{\tau_m\}_{m=1}^{M}$ over visible scene points. LongDPM addresses the scalability issue by processing the video in overlapping chunks and restoring cross-chunk consistency through overlap-aware geometric alignment and dynamic trajectory association.

\subsection{Static-Aware Overlap Abstraction}
\paragraph{Sequence Chunking.} 
To maintain a bounded memory footprint, we partition the video into $K$ overlapping chunks $\{\mathcal{X}_k\}_{k=1}^K$. Each chunk $\mathcal{X}_k = \{I_t\}_{t=s_k}^{e_k}$ has a maximum length of $L$ frames and an overlap of $O$ frames with its successor, such that $s_{k+1} = e_k - O + 1$. A frozen local dynamic reconstructor (e.g., V-DPM) is applied independently to each chunk, yielding local pointmaps $\mathbf{p}^{(k)}_{\tau,u}$, confidence maps $c^{(k)}_{\tau,u}$, and camera poses $\mathbf{E}^{(k)}_\tau$ for frame $\tau$ and pixel $u$. This chunk-and-align design follows recent long-sequence feed-forward reconstruction methods; however, in dynamic videos, moving foregrounds and boundary occlusions frequently dominate overlap regions, rendering naive averaging or direct pointmap stitching unreliable.

Since each chunk is reconstructed in its own relative coordinate system, the shared overlap $\Omega_{ij}$ between adjacent chunks $i$ and $j$ serves as the primary evidence for estimating their relative transformation. This stage not only ensures computational tractability but also distills each overlap into a compact set of geometrically reliable constraints that remain robust to dynamic motion and viewpoint changes. In particular, we preserve rigid registration evidence while deferring motion-rich regions to a later non-rigid reasoning stage. Such separation is essential because, unlike static scenes, overlap evidence in dynamic videos is inherently heterogeneous: certain pixels are reliable for rigid alignment, whereas others primarily support temporal identity association.

\paragraph{Rigid Anchor Selection.}

A naive registration of overlapping pointmaps is prone to failure in dynamic scenes, since moving objects violate the rigid transform assumption. To address this, we abstract the overlap region into two complementary sets: \textit{static anchors} $\mathcal{A}^{\text{stat}}_{ij}$ and \textit{dynamic supports} $\mathcal{A}^{\text{dyn}}_{ij}$. The core idea is to select overlap points that are both temporally stable and highly confident, rather than using all pixels equally.

Following the principle of temporal invariance, the static anchor set is defined as:
\begin{equation}
\mathcal{A}^{\text{stat}}_{ij} = \left\{ u \mid \bar{c}_u^{(k)} > \gamma_c, \,\, \max_{\tau, \tau' \in \Omega_{ij}} \| \mathbf{p}^{(k)}_{\tau,u} - \mathbf{p}^{(k)}_{\tau',u} \|_2 < \gamma_{\text{stat}} \right\},
\end{equation}
where $\bar{c}_u^{(k)}$ denotes the mean confidence across the overlap and $\gamma_{\text{stat}}$ is a rigidity threshold. The remaining high-confidence pixels that fail the rigidity test form the dynamic support set $\mathcal{A}^{\text{dyn}}_{ij}$.

This filtering ensures that only geometrically reliable and temporally stationary points contribute to rigid alignment. By suppressing dynamic foreground regions—which may dominate the image-space overlap but provide inconsistent 3D cues—we obtain a cleaner rigid registration signal while preserving valuable motion evidence for subsequent dynamic reasoning.

\paragraph{Pairwise Transformation.}

Given the static anchors, we estimate the relative similarity transform $\widehat{\mathcal{T}}_{ij} = (\mathbf{R}, \mathbf{t}, s)$ between adjacent chunks by minimizing the confidence-weighted Procrustes loss:
\begin{equation}
\mathcal{L}^{\text{ovl}}_{ij} = \sum_{\tau \in \Omega_{ij}} \sum_{u \in \mathcal{A}^{\text{stat}}_{ij}} \sqrt{c^{(i)}_{\tau,u} c^{(j)}_{\tau,u}} \left\| s\mathbf{R} \mathbf{p}^{(j)}_{\tau,u} + \mathbf{t} - \mathbf{p}^{(i)}_{\tau,u} \right\|_2^2.
\end{equation}
Points in $\mathcal{A}^{\text{dyn}}_{ij}$ are excluded from this optimization to prevent foreground motion from corrupting the camera trajectory. The confidence weighting further biases the estimate toward regions that are stable and reliable in both chunks, which is particularly important under partial occlusions or limited background structure. Overall, this module effectively converts dense but noisy local predictions into sparse, high-quality geometric constraints for robust chunk-to-chunk registration.

To handle real-world sequences where overlap evidence may be weak, our implementation adopts a practical fallback strategy: when sufficient reliable dynamic correspondences become available after later association, they are fed back to jointly refine the overlap transform together with camera centers; otherwise, the method falls back to static-anchor alignment and, if necessary, to the chunk camera poses predicted by the local reconstructor. This hierarchical approach ensures that the strongest available evidence is prioritized while maintaining stability when overlap support is limited.
\subsection{Motion-Aware Trajectory Association}

Although chunk registration reduces coordinate drift, the recovered motion remains fragmented into chunk-local tracklets. We therefore formulate long-range dynamic reconstruction as a \emph{tracklet association} problem over neighboring overlaps. Restricting association to adjacent chunks is critical: after alignment, the overlap region is the only temporal interval in which two candidate tracklets from adjacent chunks can be directly compared in a shared 3D coordinate frame, providing the most reliable cue for identity continuity. This formulation differs from query-based trackers such as TAPIR or CoTracker~\cite{doersch2023tapir,karaev2025cotracker3,zhang2025tapip3d}, which typically propagate identities from a designated query frame or support window. In contrast, our goal is to reconnect chunk-local dynamic evidence \emph{after} overlap-based geometric registration, naturally posing the correspondence problem in the aligned 3D space rather than in image coordinates.

\paragraph{Tracklet construction.}

For two temporally adjacent aligned chunks \(\mathcal{X}_i\) and \(\mathcal{X}_j\) with overlap \(\Omega_{ij}\), we reshape the overlap pointmaps into per-pixel 3D tracklets and retain only informative dynamic candidates. Specifically, we filter out tracklets with low mean confidence over \(\Omega_{ij}\) or negligible temporal displacement, thereby removing unreliable noise and quasi-static background points. The retained sets are denoted by \(\mathcal{D}_i\) and \(\mathcal{D}_j\), where each \(d \in \mathcal{D}\) represents a short 3D trajectory segment in the shared coordinate frame.

This filtering step is essential because dense 4D predictors generate many formally valid trajectories that are uninformative for identity association, such as tiny residual motions on the background, uncertain points near disocclusions, or transient outliers near chunk boundaries. Additionally, we subsample the confident dynamic seeds to maintain computational tractability while preserving sufficient spatial coverage. This design aligns with practical observations in modern tracking systems: although dense candidate generation is beneficial, assignment should operate on a carefully filtered subset to avoid spurious matches in crowded or ambiguous scenes.

\paragraph{Multi-cue association cost.}

We first prune the candidate space via endpoint gating: a pair \((d_a, d_b)\), with \(d_a \in \mathcal{D}_i\) and \(d_b \in \mathcal{D}_j\), is considered only if the Euclidean distance between their boundary 3D positions is below a radius \(\gamma_p\). In practice, this is implemented as a small nearest-neighbor search around the terminal position of each tracklet, which efficiently eliminates most implausible pairings.

The remaining candidates are then scored using a unified multi-cue cost
\begin{equation}
\mathcal{C}(d_a,d_b) = \sum_{k \in \{\mathrm{traj},\mathrm{vel},\mathrm{dir}\}} \lambda_k \mathcal{L}_k(d_a,d_b),
\end{equation}
where \(\mathcal{L}_{\mathrm{traj}}\) measures the normalized 3D trajectory discrepancy over the overlap \(\Omega_{ij}\), while \(\mathcal{L}_{\mathrm{vel}}\) and \(\mathcal{L}_{\mathrm{dir}}\) penalize inconsistencies in motion magnitude and direction, respectively. Pairs with excessive trajectory error or directional mismatch are discarded prior to final assignment.

This multi-cue design is motivated by the short length of overlap windows, where geometry alone can be ambiguous. While \(\mathcal{L}_{\mathrm{traj}}\) checks spatial proximity of the trajectories, nearby but distinct objects may still produce similar paths over a few frames. The velocity and direction terms provide critical derivative cues to reject dynamically inconsistent pairs (e.g., crossing trajectories moving in opposite directions), yielding a compact yet robust cost that leverages both geometric consistency and local motion dynamics to disambiguate short overlaps.

\paragraph{Identity-preserving assignment.}

Given the sparse pairwise cost matrix, we solve tracklet linkage as a one-to-one bipartite matching problem:
\begin{equation}
\mathcal{M}_{ij} = \arg\min_{\mathbf{Z}\in\{0,1\}^{|\mathcal{D}_i|\times|\mathcal{D}_j|}} \sum_{a,b} Z_{ab}\,\mathcal{C}(d_a,d_b).
\end{equation}
We employ the Hungarian algorithm with strict one-to-one row and column constraints, incorporating dummy nodes or a maximum-cost threshold to permit unmatched tracklets.

This formulation is critical for preventing identity collapse in cases of spatial clustering or uniform motion, while maintaining temporal continuity without duplicate assignments. Equally important is what the module deliberately does \emph{not} enforce: unmatched tracklets are retained as independent fragments. This design naturally accommodates object entry, exit, and unsupported re-appearances without relying on first-frame queries or full-sequence recurrent memory. Such open-world behavior is particularly valuable for long videos, where objects may appear late, disappear for extended periods, or re-enter under different viewpoints. By avoiding forced continuity when evidence is weak, the module preserves robustness and allows later observations to establish new identities when justified.
\subsection{Trajectory-Guided Dynamic Fusion}

Given the identity-consistent correspondences \(\mathcal{M}_{ij}\), we turn trajectory-level association into geometric constraints for dynamic fusion. Rather than treating chunk merging as a passive concatenation step, we refine the overlap transform using matched dynamic trajectories and then reconstruct continuity at the chunk boundary. This feedback loop is a key distinction from pipelines that stop once identities have been linked: in LongDPM, successful association becomes additional geometric evidence that can correct residual misalignment left by static-only registration.

\paragraph{Refined overlap alignment.}
For two adjacent chunks \(\mathcal{X}_i\) and \(\mathcal{X}_j\), each matched pair \((d_a,d_b)\in\mathcal{M}_{ij}\) provides a set of dynamic 3D correspondences over the overlap \(\Omega_{ij}\). We estimate the refined relative transform \(\widehat{\mathcal{T}}_{ij}=(\mathbf{R},\mathbf{t})\) by minimizing
\begin{equation}
\mathcal{L}_{\mathrm{ref}}
=
\mathcal{L}_{\mathrm{track}}
+
\lambda_{\mathrm{cam}}\mathcal{L}_{\mathrm{cam}},
\end{equation}
where
\begin{equation}
\mathcal{L}_{\mathrm{track}}
=
\sum_{(a,b)\in\mathcal{M}_{ij}}
\sum_{\tau\in\Omega_{ij}}
w_{ab,\tau}
\left\|
\mathbf{R}\mathbf{x}^{(j)}_{\tau}(d_b)+\mathbf{t}-\mathbf{x}^{(i)}_{\tau}(d_a)
\right\|_2^2,
\end{equation}
and
\begin{equation}
\mathcal{L}_{\mathrm{cam}}
=
\sum_{\tau\in\Omega_{ij}}
\left\|
\mathbf{R}\mathbf{c}^{(j)}_{\tau}+\mathbf{t}-\mathbf{c}^{(i)}_{\tau}
\right\|_2^2,
\end{equation}
where \(\mathbf{c}^{(i)}_{\tau}\) and \(\mathbf{c}^{(j)}_{\tau}\) denote the camera centers of the two chunks at frame \(\tau\). This joint formulation mirrors the actual behavior of our implementation, which performs a weighted Kabsch alignment on the union of matched dynamic points and pose-derived camera centers. The track term supplies dense dynamic evidence, while the pose term prevents the refined transform from drifting toward implausible solutions under sparse, unbalanced, or noisy matches.

Using dynamic points for alignment does not assume that they are static across time. The transform \(\widehat{\mathcal{T}}_{ij}\) maps one chunk coordinate system to another; at the same timestamp \(\tau\), two chunks that observe the same physical moving point should still predict the same world-space position after this coordinate conversion. Thus matched dynamic trajectories provide valid cross-coordinate constraints at overlapping frames, even though their positions change over time. This refinement is most useful when the overlap is dominated by moving objects or contains limited low-parallax background structure. In such cases, static anchors alone may be too few to determine a reliable relative transform, and associated foreground trajectories can become the only available evidence for preventing chunk drift. The weighting further reduces the impact of outlier correspondences, so that a small number of erroneous matches does not destabilize the chunk transform.

\paragraph{Boundary continuity reconstruction.}
After estimating \(\widehat{\mathcal{T}}_{ij}\), we transform the second chunk into the coordinate system of the first and reconstruct continuity for each matched tracklet pair over a short boundary window. Let \(\tilde{\mathbf{x}}_{\tau}\) denote the refined boundary trajectory, and let \(\bar{\mathbf{x}}^{(j)}_{\tau}(d_b)=\mathbf{R}\mathbf{x}^{(j)}_{\tau}(d_b)+\mathbf{t}\) denote the aligned trajectory from the second chunk. Rather than re-optimizing the entire trajectory, we apply a local continuity update near the junction by minimizing
\begin{equation}
\begin{aligned}
\mathcal{L}_{\mathrm{bnd}}
&=
\sum_{\tau\in\mathcal{B}_{ij}}
\alpha_{\tau}
\left\|
\tilde{\mathbf{x}}_{\tau}-\mathbf{x}^{(i)}_{\tau}(d_a)
\right\|_2^2
\;+\;\\
&\quad
\beta_{\tau}
\left\|
\tilde{\mathbf{x}}_{\tau}-\bar{\mathbf{x}}^{(j)}_{\tau}(d_b)
\right\|_2^2
\;+\; 
\lambda_{\mathrm{sm}}
\left\|
\tilde{\mathbf{x}}_{\tau}-\tilde{\mathbf{x}}_{\tau-1}
\right\|_2^2,
\end{aligned}
\end{equation}
where \(\mathcal{B}_{ij}\) is a short temporal window centered on the chunk boundary, and \(\alpha_{\tau},\beta_{\tau}\) balance fidelity to the two aligned tracklets. We set these weights to transition smoothly from the first chunk to the second chunk near the junction, so that the tail of \(d_a\) dominates before the boundary while the aligned head of \(d_b\) dominates after it. The smoothness weight \(\lambda_{\mathrm{sm}}\) controls the trade-off between preserving the local predictor outputs and suppressing abrupt position or velocity jumps. In practice, this is implemented as a short-horizon smoothing/blending step over the tail of the first segment toward the aligned head of the second, which is sufficient to remove visible boundary artifacts.

Restricting this reconstruction to a short boundary window is important for two reasons. First, it directly targets the artifact introduced by independent chunk inference, namely a discontinuity at the overlap junction. Second, it avoids oversmoothing interior motion that is already reliable within each chunk, preserving the short-range fidelity of the local predictor. The non-overlapping suffix of the second chunk is then appended after alignment, producing trajectories that are both identity-consistent and temporally smooth.

\paragraph{Sequence-level fusion.}
Applying this procedure recursively over adjacent chunk pairs produces a unified dynamic reconstruction in a shared world frame. Each merge step depends only on neighboring overlaps, which keeps memory bounded while progressively recovering sequence-level geometric consistency and cross-boundary motion continuity. The final representation combines aligned dense pointmaps with trajectories \(\mathcal{T}\) that are extended when neighboring overlaps provide sufficient evidence, while unsupported fragments remain separate.

Viewed as a whole, this module links geometry and tracking. The first stage extracts rigid overlap evidence, the second stage recovers dynamic identities in the aligned space, and the present stage feeds those identities back into geometric fusion. This cross-chunk feedback pattern allows LongDPM to inherit the short-context reconstruction quality of feed-forward 4D models while extending them to longer videos without requiring whole-sequence joint inference.

\section{Experiments}

We evaluate LongDPM along three axes used by recent dynamic point-map methods: two-view 4D reconstruction, dense multi-frame tracking, and camera pose estimation. Tables~\ref{tab:main_results}--\ref{tab:tracking_epe} verify that chunked inference preserves local accuracy, while Table~\ref{tab:ablation} isolates the effect of the proposed sequence-level components.

\subsection{Implementation Details}

\paragraph{Setup.}
LongDPM employs a frozen local dynamic point-map predictor and applies the proposed overlap-aware reasoning exclusively across chunk boundaries. For long-sequence experiments, we adopt a chunk length of $C=16$ frames, an overlap of $O=4$ frames, and an input resolution of $518\times518$ pixels, unless otherwise specified in the corresponding tables. The overlap length is chosen as a practical trade-off: it provides sufficient frames to reliably estimate local velocity and motion direction for cross-chunk association, while keeping the computational redundancy of overlapping chunk inference minimal. All experiments are conducted on a single NVIDIA GeForce RTX 5090 GPU with 36GB memory. We compare LongDPM against DPM~\cite{sucar2025dynamic}, V-DPM~\cite{sucar2026v}, St4RTrack~\cite{feng2025st4rtrack}, TraceAnything~\cite{liu2025trace}, and standard pose estimation baselines.

\begin{figure}[t]
    \raggedleft
    \includegraphics[width=0.95\columnwidth]{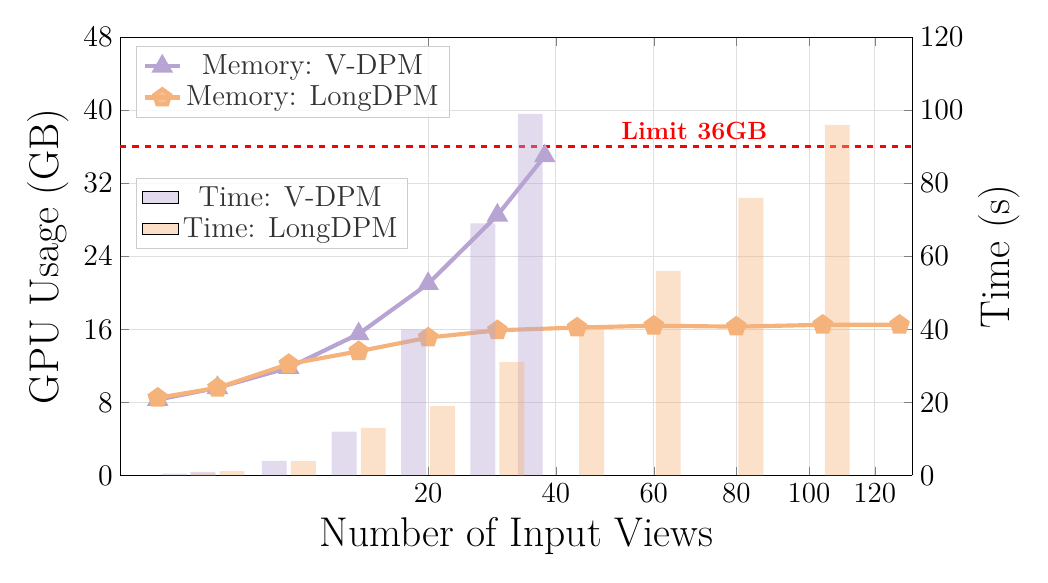}
    \caption{Memory and runtime scalability as the number of input views increases.}
    \label{fig:table}
\end{figure}

\paragraph{Test Datasets and Metrics.}
We follow the standard evaluation protocols established by recent dynamic point-map methods. For 4D reconstruction and dense tracking, we evaluate on PointOdyssey~\cite{zheng2023pointodyssey}, Kubric-F, Kubric-G~\cite{greff2021kubric}, and Waymo~\cite{sun2020scalability}, reporting 3D end-point error (EPE) in the world coordinate frame of the first view. For camera pose estimation, we use Sintel~\cite{butler2012naturalistic} and TUM-dynamics~\cite{sturm2012benchmark}, with metrics including Absolute Trajectory Error (ATE), Relative Pose Error (RPE) in translation, and RPE in rotation.

\subsection{Comparison with Prior Work}

\begin{figure*}[t]
    \centering
    \includegraphics[width=\textwidth]{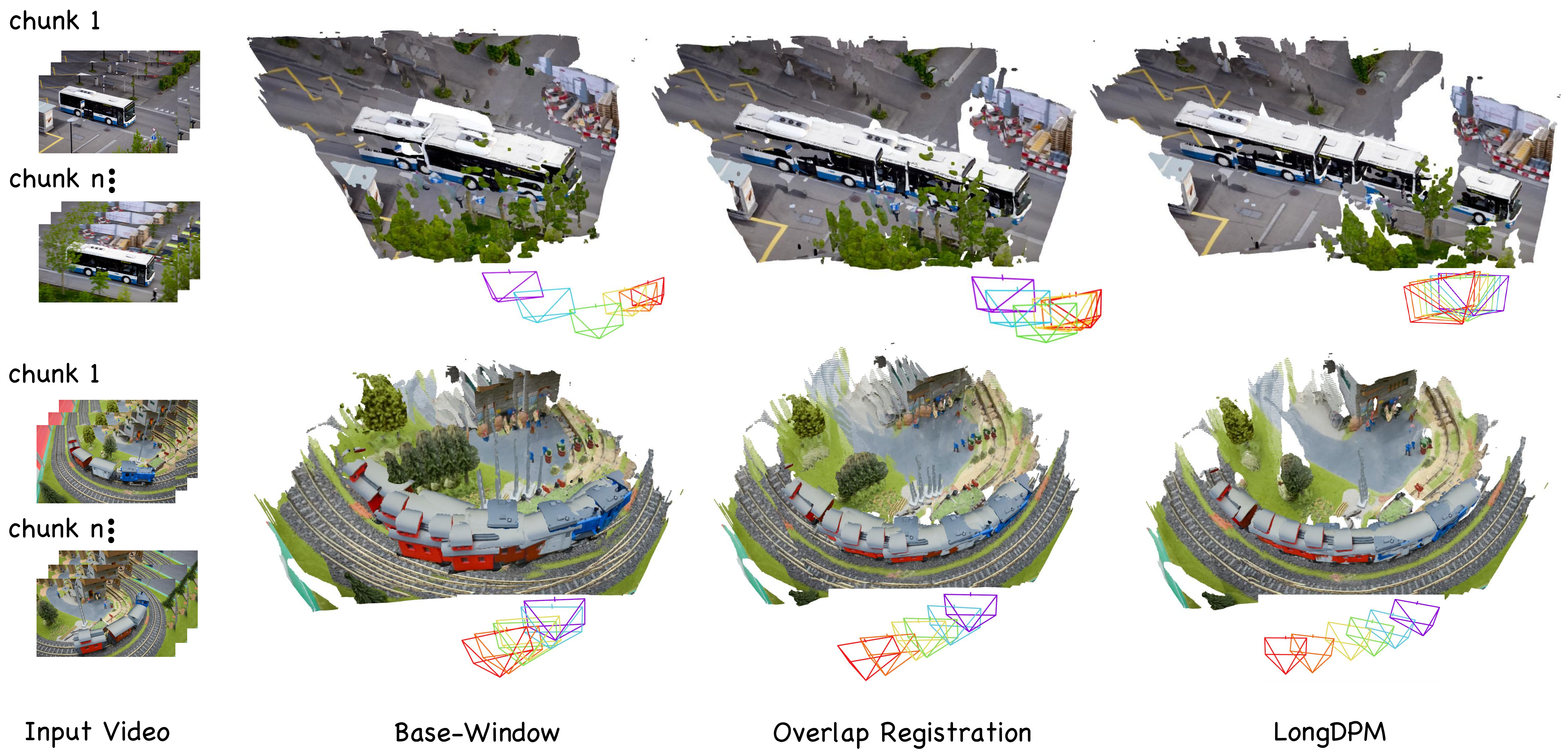}
    \caption{
    \textbf{Cross-chunk reconstruction comparison.}
    Base-window inference reconstructs video chunks independently and produces inconsistent geometry and camera poses across chunks. Overlap registration reduces this misalignment by aligning adjacent chunks through shared frames, but can still be unstable under dynamic motion and weak overlap. LongDPM uses static-aware overlap abstraction and dynamic association to recover a more coherent long-range reconstruction in a shared coordinate system.
    }
    \label{fig:comparison}
\end{figure*}
We compare LongDPM with prior work under the protocols summarized above, using a chunk size of 16 for LongDPM unless otherwise noted, and additionally include qualitative cross-chunk reconstruction. Figure~\ref{fig:table} reports memory and runtime scalability as the number of input views increases. Since V-DPM performs whole-sequence inference, its GPU memory grows rapidly with sequence length and exceeds the 36GB limit for long inputs. LongDPM keeps memory nearly bounded around the fixed chunk size while supporting substantially longer videos, with only moderate runtime growth from processing additional chunks.

\paragraph{Two-view dynamic reconstruction.}
We first evaluate LongDPM on the standard two-view dynamic 3D reconstruction protocol used by DPM and V-DPM. Given two sampled video frames, the model reconstructs four point maps, corresponding to the 3D positions of points from each image at both timestamps. Table~\ref{tab:main_results} reports EPE under temporal margins 8 and 16 in the first-view world coordinate frame, so the metric jointly reflects reconstruction, camera estimation, and temporal correspondence. LongDPM improves over prior feed-forward methods on nearly all entries and consistently outperforms V-DPM on PointOdyssey, Kubric-F, and Kubric-G. This indicates that the chunked formulation can preserve local point-map accuracy while maintaining a stable coordinate frame. On Waymo, the final point-map entry remains slightly better for V-DPM, suggesting that this rigid driving benchmark benefits less from dynamic cross-chunk reasoning.

\begin{table*}[t]
\centering
\small
\setlength{\tabcolsep}{4pt}
\renewcommand{\arraystretch}{1.15}
\begin{adjustbox}{width=\textwidth,center}
\begin{tabular}{lcccccccccccccccc}
\toprule
\multirow{2}{*}{\textbf{Method}} 
& \multicolumn{4}{c}{\textbf{PointOdyssey}}
& \multicolumn{4}{c}{\textbf{Kubric-F}}
& \multicolumn{4}{c}{\textbf{Kubric-G}}
& \multicolumn{4}{c}{\textbf{Waymo}} \\
\cmidrule(lr){2-5} \cmidrule(lr){6-9} \cmidrule(lr){10-13} \cmidrule(lr){14-17}
& $P_0(t_0)$ & $P_0(t_1)$ & $P_1(t_0)$ & $P_1(t_1)$
& $P_0(t_0)$ & $P_0(t_1)$ & $P_1(t_0)$ & $P_1(t_1)$
& $P_0(t_0)$ & $P_0(t_1)$ & $P_1(t_0)$ & $P_1(t_1)$
& $P_0(t_0)$ & $P_0(t_1)$ & $P_1(t_0)$ & $P_1(t_1)$ \\
\midrule
\multicolumn{17}{c}{\textbf{Margin: 8}} \\
\midrule
St4RTrack
& --- & 0.143 & --- & 0.146
& --- & 0.163 & --- & 0.059
& --- & 0.193 & --- & 0.113
& --- & 0.232 & --- & 0.261 \\
TraceAnything
& 0.151 & 0.156 & 0.166 & 0.165
& 0.082 & 0.115 & 0.127 & 0.091
& 0.094 & 0.139 & 0.154 & 0.130
& 0.188 & 0.192 & 0.235 & 0.235 \\
DPM
& 0.101 & 0.103 & 0.103 & 0.104
& 0.030 & 0.050 & 0.044 & 0.039
& 0.041 & 0.068 & 0.065 & 0.051
& 0.085 & 0.085 & 0.083 & 0.084 \\
V-DPM
& 0.029 & 0.031 & 0.032 & 0.030
& 0.017 & 0.039 & 0.033 & 0.025
& 0.022 & 0.049 & 0.045 & 0.029
& 0.065 & 0.067 & 0.065 & \textbf{0.064} \\
\textbf{LongDPM (Chunk Size=8)}
& \textbf{0.024} & \textbf{0.027} & \textbf{0.028} & \textbf{0.026}
& \textbf{0.013} & \textbf{0.031} & \textbf{0.027} & \textbf{0.020}
& \textbf{0.017} & \textbf{0.043} & \textbf{0.038} & \textbf{0.024}
& \textbf{0.058} & \textbf{0.061} & \textbf{0.059} & 0.065 \\
\midrule
\multicolumn{17}{c}{\textbf{Margin: 16}} \\
\midrule
St4RTrack
& --- & 0.142 & --- & 0.145
& --- & 0.180 & --- & 0.075
& --- & 0.215 & --- & 0.137
& --- & 0.239 & --- & 0.296 \\
TraceAnything
& 0.148 & 0.158 & 0.171 & 0.170
& 0.101 & 0.162 & 0.180 & 0.121
& 0.111 & 0.190 & 0.211 & 0.178
& 0.226 & 0.232 & 0.304 & 0.306 \\
DPM
& --- & --- & --- & ---
& --- & --- & --- & ---
& --- & --- & --- & ---
& --- & --- & --- & --- \\
V-DPM
& 0.030 & 0.033 & 0.034 & 0.031
& 0.019 & 0.058 & 0.049 & 0.036
& 0.025 & 0.073 & 0.067 & 0.041
& 0.067 & 0.070 & 0.068 & \textbf{0.066} \\
\textbf{LongDPM (Chunk Size=8)}
& \textbf{0.025} & \textbf{0.029} & \textbf{0.030} & \textbf{0.027}
& \textbf{0.015} & \textbf{0.047} & \textbf{0.040} & \textbf{0.029}
& \textbf{0.020} & \textbf{0.064} & \textbf{0.057} & \textbf{0.034}
& \textbf{0.061} & \textbf{0.065} & \textbf{0.063} & 0.067 \\
\bottomrule
\end{tabular}
\end{adjustbox}
\caption{\textbf{Two-view tracking EPE}, reported for four point clouds (one for each image and time frame). LongDPM is evaluated under the same protocol with margins 8 and 16.}
\label{tab:main_results}
\end{table*}

\begin{table}[t]
\centering
\small
\setlength{\tabcolsep}{3pt}
\begin{adjustbox}{width=\columnwidth,center}
\begin{tabular}{lcccccc}
\toprule
Method & \multicolumn{3}{c}{Sintel} & \multicolumn{3}{c}{TUM-dynamics} \\
\cmidrule(lr){2-4}\cmidrule(lr){5-7}
 & ATE $\downarrow$ & \makecell{RPE trans $\downarrow$} & \makecell{RPE rot $\downarrow$} & ATE $\downarrow$ & \makecell{RPE trans $\downarrow$} & \makecell{RPE rot $\downarrow$} \\
\midrule
Robust-CVD & 0.360 & 0.154 & 3.443 & 0.189 & 0.071 & 3.681 \\
CasualSAM & 0.141 & \textbf{0.035} & 0.615 & 0.045 & 0.020 & 0.841 \\
DUSt3R & 0.417 & 0.250 & 5.796 & 0.127 & 0.062 & 3.099 \\
MonST3R & 0.108 & 0.042 & 0.732 & 0.074 & 0.019 & 0.905 \\
DPM & -- & -- & -- & 0.056 & 0.014 & 0.836 \\
$\pi^3$ & \textbf{0.074} & 0.040 & 0.282 & 0.014 & 0.009 & \textbf{0.312} \\
V-DPM & 0.105 & 0.048 & 0.670 & 0.057 & 0.017 & 0.340 \\
LongDPM (ours) & 0.078 & 0.037 & \textbf{0.274} & \textbf{0.012} & \textbf{0.008} & 0.318 \\
\bottomrule
\end{tabular}
\end{adjustbox}
\caption{Camera pose evaluation on Sintel and TUM-dynamics. Bold indicates the best performance.}
\label{tab:pose_results}
\end{table}

\paragraph{Camera pose estimation.}
Next, we evaluate camera pose estimation on Sintel and TUM-dynamics following the protocols of MonST3R and V-DPM. Table~\ref{tab:pose_results} reports ATE, RPE translation, and RPE rotation, which test whether LongDPM can maintain reliable camera geometry even though inference is performed in overlapping chunks. LongDPM achieves the best TUM-dynamics ATE and RPE translation, as well as the best Sintel RPE rotation, while remaining close to $\pi^3$ on the remaining metrics. These results show that the proposed overlap abstraction is not only useful for dynamic tracking, but also provides a stable rigid backbone for camera alignment. The strong performance on TUM-dynamics is especially relevant because moving foreground objects can otherwise corrupt pose estimation if they are not separated from reliable static evidence.

\begin{table}[t]
\centering
\small
\setlength{\tabcolsep}{3pt}
\resizebox{0.85\columnwidth}{!}{
\begin{tabular}{lcccc}
\toprule
Method & PointOdyssey & Kubric-F & Kubric-G & Waymo \\
\midrule
St4RTrack     & 0.137 & 0.153 & 0.201 & 0.167 \\
TraceAnything & 0.152 & 0.107 & 0.126 & 0.119 \\
DPM           & 0.114 & 0.088 & 0.109 & 0.103 \\
V-DPM & 0.032 & 0.027 & 0.035 & \textbf{0.042} \\
\textbf{LongDPM} & \textbf{0.025} & \textbf{0.021} & \textbf{0.029} & 0.044 \\
\bottomrule
\end{tabular}
}
\caption{\textbf{Tracking EPE error} reported for 10-frame snippets, evaluating dense tracks of all pixels in the first frame.}
\label{tab:tracking_epe}
\end{table}

\paragraph{Dense 3D tracking.}
Finally, we consider dense 3D tracking over 10-frame snippets, where all pixels in the first frame are tracked and evaluated by average EPE over their predicted 3D trajectories. As shown in Table~\ref{tab:tracking_epe}, LongDPM improves over V-DPM on PointOdyssey, Kubric-F, and Kubric-G, reducing EPE from 0.032 to 0.025, 0.027 to 0.021, and 0.035 to 0.029, respectively. This benchmark tests whether dynamic identities remain consistent under the same local prediction protocol used by prior work. The larger gain compared with the pose-only benchmark suggests that LongDPM's main advantage comes from maintaining dynamic correspondences across chunk boundaries, rather than from camera estimation alone. Waymo again shows a small gap in favor of V-DPM, consistent with its stronger rigid driving structure.

\paragraph{Visualization Results.}
Figure~\ref{fig:comparison} provides a qualitative comparison of cross-chunk reconstruction. While base-window inference produces plausible local geometry but suffers from inconsistent coordinates and camera poses across chunks, and simple overlap registration still struggles with dynamic motion and weak overlap, LongDPM achieves significantly more coherent long-range geometry and temporally consistent trajectories. By leveraging static-aware overlap abstraction and dynamic trajectory association, our method effectively maintains both geometric consistency and motion continuity across chunk boundaries.

\subsection{Ablation Study}

The ablation study serves as a key experiment for evaluating LongDPM, as it isolates the contribution of the proposed cross-chunk modules while controlling for backbone and protocol differences. All variants employ the same frozen local predictor, chunk length, overlap size, input resolution, and evaluation protocol; only the cross-chunk components differ. Specifically, we compare three variants: (i) the base-window variant, which directly concatenates independent chunk predictions; (ii) the overlap-abstraction variant, which adds static-aware overlap registration; and (iii) the full model, which further incorporates dynamic identity association and trajectory-guided fusion.

\begin{table}[t]
\centering
\small
\setlength{\tabcolsep}{3pt}
\renewcommand{\arraystretch}{1.12}
\begin{adjustbox}{width=\columnwidth,center}
\begin{tabular}{lcccc}
\toprule
Variant & PointOdyssey & Kubric-F & Kubric-G & Waymo \\
\midrule
Base-window & 0.039 & 0.034 & 0.045 & 0.057 \\
+ overlap abstraction & 0.033 & 0.028 & 0.038 & 0.050 \\
\textbf{+ dynamic association} & \textbf{0.025} & \textbf{0.021} & \textbf{0.029} & \textbf{0.044} \\
\bottomrule
\end{tabular}
\end{adjustbox}
\caption{\textbf{Component ablation on 10-frame dense tracking EPE.} We progressively add overlap abstraction and dynamic association to the same chunked local predictor.}
\label{tab:ablation}
\end{table}

\paragraph{Effect of overlap abstraction.}
The base-window variant yields the highest error across all datasets, indicating that independently reconstructed chunks cannot be naively concatenated due to coordinate misalignment. Incorporating static-aware overlap abstraction reduces the average EPE from 0.044 to 0.037. This improvement demonstrates the importance of filtering dynamic foregrounds and using confidence-weighted, rigidity-constrained static anchors for robust registration. The result validates our design choice of decoupling rigid spatial alignment from subsequent motion-aware reasoning.

\paragraph{Effect of dynamic association and refinement.}
Adding geometry-conditioned dynamic association and trajectory-guided fusion further reduces the average EPE from 0.037 to 0.030, achieving the best performance on every dataset in Table~\ref{tab:ablation}. Notably, this stage goes beyond simple tracking post-processing: the matched dynamic tracklets provide additional geometric constraints at chunk boundaries, enabling the refinement step to correct residual misalignment that static anchors alone cannot resolve. This explains the larger gains observed on dynamic synthetic benchmarks, where foreground motion, once properly associated, contributes valuable geometric information for improving overall consistency.

\section{Conclusion}

We present LongDPM, a framework for long-range monocular dynamic reconstruction that extends short-context feed-forward predictors to longer videos via overlap-aware cross-chunk reasoning. By decoupling local reconstruction from global consistency, LongDPM maintains inference memory bounded by chunk length while producing dense geometry, camera motion, and coherent long-range 3D trajectories in a unified coordinate system. The proposed static-aware overlap abstraction, geometry-conditioned identity association, and trajectory-guided dynamic fusion effectively improve cross-chunk alignment and motion continuity. Experiments on dynamic reconstruction, dense 3D tracking, and camera pose estimation show that LongDPM preserves local accuracy while achieving superior long-range consistency compared to prior methods. Future work includes adaptive chunk scheduling, more reliable overlap estimation for challenging cases such as unreliable geometry, extremely fast motion, and severe occlusions, as well as tighter joint reasoning between geometric fusion and dynamic identity association.

\subsection{Acknowledgement}
This project is supported by the Zhejiang Provincial Natural Science Foundation of China (No. LQN25F020019) and the Key Laboratory of Data Science and Intelligence Education (Hainan Normal University), Ministry of Education (No. DSIE202403).

\bibliography{main}
\clearpage

\end{document}